\documentclass[11pt,letterpaper]{article}
\usepackage[hyperref]{emnlp-ijcnlp-2019}

\usepackage{amsmath}
\usepackage{amssymb}
\usepackage{graphicx}
\usepackage{booktabs}
\usepackage{url}
\usepackage[linguistics]{forest}
\usepackage{tikz}
\usetikzlibrary{calc}

\aclfinalcopy 

\newcommand{\figref}[1]{Figure~\ref{fig:#1}}
\newcommand{\secref}[1]{Section~\ref{sec:#1}}
\newcommand{\tabref}[1]{Table~\ref{tab:#1}}
\renewcommand{\eqref}[1]{Equation~\ref{eq:#1}}

\newcommand{\F}{\mathcal{F}}
\renewcommand{\H}{\mathcal{H}}
\newcommand{\C}{\mathcal{C}}
\renewcommand{\L}{\mathcal{L}}

\usepackage [autostyle, english = american]{csquotes}
\MakeOuterQuote{"}

\title{Leveraging External Knowledge for Out-Of-Vocabulary Entity Labeling}
\author{
  Adrian de Wynter \\
  Amazon Alexa \\
  {\tt dwynter@amazon.com} \\\And
  Lambert Mathias \\
  Amazon Alexa \\
  {\tt mathiasl@amazon.com}
}
\date{}

\begin{document}
\maketitle
\begin{abstract}
  Dealing with previously unseen slots is a challenging problem in a real-world multi-domain dialogue state tracking task. Other approaches rely on predefined mappings to generate candidate slot keys, as well as their associated values. This, however, may fail when the key, the value, or both, are not seen during training. To address this problem we introduce a neural network that leverages external knowledge bases (KBs) to better classify out-of-vocabulary slot keys and values. This network projects the slot into an attribute space derived from the KB, and, by leveraging similarities in this space, we propose candidate slot keys and values to the dialogue state tracker. 
We provide extensive experiments that demonstrate that our stratagem can improve upon a previous approach, which relies on predefined candidate mappings. 
In particular, we evaluate this approach by training a state-of-the-art model with candidates generated from our network, and obtained relative increases of 57.7\% and 82.7\% in F1 score and accuracy, respectively, for the aforementioned model, when compared to the current candidate generation strategy.

\end{abstract}
\section{Introduction}\label{sec:introduction}

In the Dialogue State Tracking task, the Slot Carryover (SC) paradigm \cite{naik2018contextualSC} involves tracking a key-value pair (a slot) throughout the dialogue. At every turn, a decision is made to whether to carry the slot over to the next turn.
The values that each slot can take are established \emph{a priori} in a hierarchical schema where the slot's value has a key assigned to it, which is itself associated with a domain. This is used to generate the candidates to be tracked, but slot-to-slot mappings can be challenging, as the keys and values could have been unseen during training \cite{xu-hu-2018-end}.

\begin{table}[ht]
\centering
\small
\begin{tabular}{|c|c|c|c|c|c|}\hline
Domain & Slot Value & Slot Key \\\hline\hline
Weather& Boston & City\_Weather \\\hline
Search & Boston & Search\_Location \\\hline
\end{tabular}
\caption{An example of a cross-domain slot. Note that there is no semantic difference in the word "Boston" across domains--"Boston" \emph{is a} city--and is only a change in naming conventions.}
\label{tab:slot_example}
\end{table}

This mapping relies on context to resolve ambiguity from values and keys. When a slot is mapped across domains, the given key is translated to its new representation on the target domain. See table \tabref{slot_example} for an example.
Two insights can be taken from this: first, the meaning of the key is preserved, and is only a change in naming conventions. Second, the key-value relationship could be seen--loosely speaking--as a type of hypernymic property. These semantic properties, along with the challenge presented by out-of-vocabulary (OOV) keys and values, motivates our work on this paper.

We surmise that it is possible to leverage knowledge bases (KBs) to build a model which automates the slot candidate generation, and circumvents the OOV problem. Our model collects several prospects, and determines the closest match with respect to the given context. We do not make assumptions about the geometry of the input spaces, and we opt instead to learn the metric as a secondary task, through the polynomial approximation to a Bregman divergence \cite{Banerjee2004ClusteringWB}.

Previous work exists on inferring labels with little to no information provided during training, such as zero and few-shot learning methods \cite{Snell2017PrototypicalNF}, and hybrid approaches such as in \cite{Akata2013LabelEmbeddingFA, yogatama2015embedding, Shi2018MultiContextLE}.
Leveraging the hierarchical nature of a corpus is not a new approach either, as hypernym classification requires some sort of hierarchical assumption, for example, \cite{Murty2018HierarchicalLA, Fu2014LearningSH, Nickel2017PoincarEF}. 
Work has also been made on employing KBs to improve learning, as in \cite{Lee2015LeveragingKB,Yang2017LeveragingKB}.
Finally, Bregman-based methods have garnered popularity as both a metric and a learning algorithm. A good example is the work of Banerjee et al., \shortcite{Banerjee2004ClusteringWB} on clustering with Bregman divergences.

Our work is unique in its approach: it relies on a KB, but places emphasis on the hierarchy of the corpus to learn and measure distances. It leverages hypernyms as an implicitly hierarchical slot key namespace. It is important to note, however, that our problem is unrelated to the hypernym classification task.

The three contributions of this paper are: first, a model that is able to output slot candidates for a dialogue state tracker (DST), even when the slot's value, key, or both, are OOV. This is done by leveraging KBs to learn the key's attributes. Second, a method to approximate a generalized, hierarchical distance function--the Bregman divergence--without making assumptions about the geometry of the space we are working on. Third, an analysis of the performance of a DST in OOV contexts, under both its current hardcoded mappings, and our model. Our approach increases the performance of the DST by 26.01\% and 51.08\% in $F_1$ and accuracy, respectively, under 100\% OOV settings. 

We start out by describing our model in Section \ref{sec:model}, and then, in \secref{experiments}, we present the results obtained from the evaluation of a DST with the candidates generated by our approach. We conclude in \secref{conclusion} with a discussion of our work.

\section{Model}\label{sec:model}
\def\layersep{2cm}

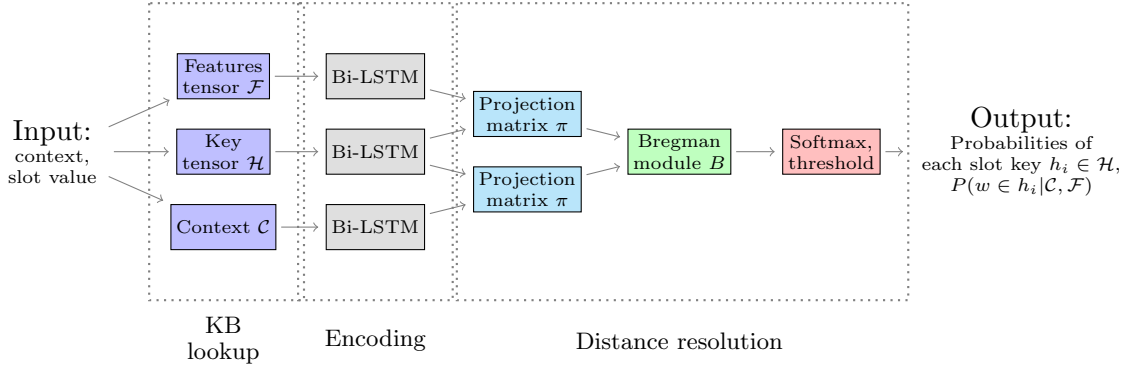
\begin{figure*}[ht!]
\centering
\begin{tikzpicture}[shorten >=1pt,->,draw=black!50, node distance=\layersep]

  \tikzstyle{every pin edge}=[<-,shorten <=1pt]
  \tikzstyle{neuron}=[circle,draw=black,fill=black!25,minimum size=17pt,inner sep=2pt, outer sep=2pt, font=\scriptsize,align=center]
  \tikzstyle{component}=[rectangle,draw=black,fill=black!25,minimum size=17pt,inner sep=2pt, outer sep=2pt, font=\scriptsize]
  \tikzstyle{input neuron}=[neuron, fill=white!0];
  \tikzstyle{output neuron}=[neuron, fill=red!50];
  \tikzstyle{hidden neuron}=[neuron, fill=blue!25];
  \tikzstyle{annot} = [text width=4em, text centered];
  \tikzstyle{box_name} = [text width=5em, text centered, font=\small];

  \node[input neuron, draw=white,outer sep=0pt] (I-1) at (-0.25,-1.5) {\normalsize{Input:} \\context, \\ slot value};

  \path[yshift=0.5cm]
  node[hidden neuron,rectangle,align=center] (H-1) at (\layersep,-1 cm) {Features \\ tensor $\F$};
  \path[yshift=0.5cm]
  node[hidden neuron,rectangle,align=center] (H-2) at (\layersep,-2 cm) {Key \\ tensor $\H$};
  \path[yshift=0.5cm]
  node[hidden neuron,rectangle,align=center] (H-3) at (\layersep,-3 cm) {Context $\C$};
  \foreach \dest in {1,...,3}
    \path (I-1) edge (H-\dest);

  \foreach \name / \y in {1,...,3}
  \path[yshift=0.5cm] node[hidden neuron,rectangle,fill=gray!25] (LSTM-\name) at (2*\layersep,-\y cm) {Bi-LSTM};
  \foreach \dest in {1,...,3}
    \path (H-\dest) edge (LSTM-\dest);

  \foreach \name / \y in {1,...,2}
    \path[yshift=0.0cm]
    node[component, fill=cyan!25,align=center] (Projection-\name) at (3*\layersep,-\y cm) {Projection \\matrix $\pi$};

  \path (LSTM-1) edge (Projection-1);
  \path (LSTM-2) edge (Projection-1);
  \path (LSTM-2) edge (Projection-2);
  \path (LSTM-3) edge (Projection-2);

  \node[component, fill=green!25,align=center] (bregman) at (4*\layersep, -1.5){Bregman \\module $B$};
  \foreach \source in {1,...,2}
    \path (Projection-\source) edge (bregman);

  \node[neuron,fill=red!25,align=center, rectangle, right of=bregman] (softmax) {Softmax, \\threshold};
  \path (bregman) edge (softmax);

  \node[input neuron,align=center, right of=softmax, draw=white,outer sep=0pt] (output) at (10.5, -1.5) {\normalsize{Output:} \\Probabilities of \\ each slot key $h_i \in \H$, \\ $P(w \in h_i \vert \C, \F)$};
  \path (softmax) edge (output);

  \draw[thick, dotted] ($(H-1.north west)+(-0.3,0.6)$)  rectangle ($(H-3.south east)+(0.3,-0.6)$);
  \draw[thick, dotted] ($(LSTM-1.north west)+(-0.3,0.6)$)  rectangle ($(LSTM-3.south east)+(0.3,-0.6)$);
  \draw[thick, dotted] ($(Projection-1.north west)+(-0.15,1.1)$) rectangle ($(softmax.south east)+(0.3,-1.6)$);

  \node[box_name,below of=LSTM-3, node distance=1.5cm](enc_text){Encoding};
  \node[box_name,below of=H-3, node distance=1.5 cm] {KB \\lookup};
  \node[box_name,below of=bregman, text width=10em, node distance=2.5cm] {Distance resolution};

\end{tikzpicture}
\caption{Architecture of our model. It computes the closest key $h_i \in \H$ for a slot value $w$, given a context $\C$ and a feature tensor $\F$. Both $\F$ and $\H$ are constructed by looking up $w$ in a set of KBs. The key-value pair is then passed in to SC for evaluation.}
\label{fig:model_main}
\end{figure*}

We aim to learn a conditional probability distribution $P(w \in h_i \vert \C, \F)$ which models the closest representative slot key $h_i \in \H$ for a slot value $w$ given both a context $\C$, and the set of possible features for all slot keys $\F$. As a secondary objective, we set the constraint that our definition of "closeness" will be relative to a learned distance $D$. 

Let $w$ be a slot value to be classified. We first construct its \emph{feature tensor} $\F$ by performing a lookup of $w$ on a set of semantically disjoint KBs, $\mathcal{K}$, such that each $\F_{ijk} \in \F$ is the $k^{th}$ set of related, yet not defining, terms to $w$, corresponding to the $j^{th}$ entry of $KB_i \in \mathcal{K}$. The assumption here is that a word is a collection of specific linguistic features \cite{Brinton2000English} and realized by its context \cite{Mikolov2013DistributedRO}. We limit the results from each $KB_i$ to $n$ entries, sorted by a relevance metric inherent to each $KB_i$, e.g., \emph{number of playbacks} in $KB_{\text{songs}}$.

As a brief example, let $w=$ "Clint Eastwood" and our KB set $\mathcal{K} = \{KB_{\text{Personalities}}\}$. $\F$ is then composed of a single set $\F_{\text{\textbf{personalities}}} = [\emph{actor}, \emph{director}, \emph{composer}, \emph{mayor}]$. Each of these fields contain their own attributes, with their own corresponding values, such as $\F_{\text{\textbf{personalities}}, \emph{mayor}, \text{location}} =$ "Camel-by-the-Sea, CA".

We then obtain the \emph{key tensor}, $\H$, by performing a reverse traversal of the Wordnet \cite{Miller1992WORDNETAL} hypernyms for the feature vectors, $\H_{i} \in \H,\; \H_{i} = WordNet(\F_{ij})$. The traversal is constrained to return only the first $m$ antecessor nodes, and we prune out the repeated entries.

Finally, we generate the \emph{context vector} $\C$, which is a flattened, tokenized version of the input context.

The three streams constructed ($\F$, $\H$, and $\C$) are flattened, embedded in GloVE \cite{pennington2014glove}, and fed into independent Bi-LSTM encoders \cite{Graves2013SpeechRW}. The independence is needed due to the fact that their inputs are assumed to belong to different semantic spaces.

\begin{equation}
  \begin{aligned}[t]
  F &= BiLSTM_F(\F) \\
  H &= BiLSTM_H(\H) \\
  C &= BiLSTM_D(\C)
  \end{aligned}
\end{equation}

Then, we extract the relationship amongst each of the encodings by projecting them onto one another.

\begin{equation}
  \begin{aligned}[t]
  \pi_F &= FH^T \\
  \pi_C &= CH^T
  \end{aligned}
\end{equation}

We split the projection matrices $\pi_C$ and $\pi_F$ in $\vert \H \vert$-sized chunks, each corresponding to one of the slot keys in the original key tensor, and evaluate the distance between $\pi_F$ and $\pi_C$.

Abstractly, the context is referring to a single entry in $\F$, and our goal is to find out which one is the closest in meaning. Hence, we require a way to measure it such that $\delta(\pi_F, \pi_C) \neq \delta(\pi_C, \pi_F)$ for some pseudometric $\delta$. Many functions could meet this requirement, but most of the more commonly used ones are instances of the Bregman divergence \cite{Banerjee2004ClusteringWB}. Choosing the right Bregman divergence for a given space is a complex problem, and is beyond the scope of this task. Therefore, we opt to learn it through a polynomial approximation of fixed degree $r$.
For this, we construct a simple feed-forward neural network $P$ of depth $r$ and output size of $1$, with a $ReLU$ unit as the final activation layer \cite{Nair2010RectifiedLU}. Our final Bregman approximator module $B$ for some $x \in \pi_C, y \in \pi_F$ is given by:

\begin{equation}
B(x, y) = P(x) - P(y) - \langle x - y, \psi(P, y, \epsilon) \rangle
\end{equation}

Where $\psi(P, y, \epsilon) = \frac{P(x + \epsilon) - P(x - \epsilon)}{2\epsilon}$ is the symmetric difference quotient estimator for the first derivative of $P$, where $y$ is the point near which to estimate the function, and $\epsilon$ a step size.

We use $B$ to extract the distances of the context to the features with respect to their given hypernyms:
\begin{equation}
D = \{B(c_i, f_i), c_i \in \pi_C, f_i \in \pi_F \}
\end{equation}

Finally, we pass in the distances into a softmax function $s(x)_i = \frac{e^{x_i}}{\Sigma_j x_j}$ to obtain their probablistic representation. 
Softmax was the preferred layer due to the fact that it tends to skew results in such a way that allow us to minimize the distance between the obtained and the expected probability distributions.

\begin{equation}
P(w \in h_i \vert \C, \F) = s(d_i), \;\; d_i \in D, h_i \in H
\end{equation}

Due to the nature of our approach and the normalization from $s(x)_i$, there are many low-confidence distances that will add noise, and could be detrimental to our training goal. To mitigate this, we implement a learnable threshold $\tau$ that allows us to prune out results:

\begin{equation}
P_{\tau}(w \in h_i \vert \C, \F) \left\{
\begin{array}{ll}
      1 & s(d_i) \geq \tau \\
      0 & else \\
\end{array}
\right.
\end{equation}

\subsection{Training}\label{sec:train}

During training, our input is of the form $(w, \C, \ell)$, where $w$ is the input slot value, along with its context $\C$ and an expected slot key $\ell$, which is just an indicator to disambiguate $w$.

We minimize a loss function $\L(y, \hat{y})$ between our predictions $y = \{P(w \in h_i \vert \C, \F) \vert h_i \in \H\}$ and our expected clusters $\hat{y}$. We construct $\hat{y}$ dynamically, by performing a fuzzy search of $\ell$ in the key tensor, indexing it, and then normalizing it: $\hat{y} = \{\hat{h_i} \vert \hat{h_i} = \frac{1}{\vert \H \vert}\mathbf{1}_\ell(h_i)\}$. The loss function is the KL divergence between our distributions:

\begin{equation}\label{eq:loss}
\L(y, \hat{y}) = - \Sigma_{i} \hat{y}_i log(\frac{\hat{y_i}}{y_i})
\end{equation}

\begin{table*}[ht]
\centering
\resizebox{\textwidth}{!}{
\begin{tabular}{|c||c|c|c|c|c|}\hline
Strategy & ${F_1}$ OOV (0\%) & Accuracy OOV (0\%) & ${F_1}$ OOV (100\%) & Accuracy OOV (100\%) \\\hline\hline
HM            & \bf{87.10} & \bf{95.12} & 32.21 & 31.57 \\\hline
MLP           & 86.77 & 94.96 & 32.21 & 31.57 \\\hline
Our approach  & 73.75 & 90.43 & \bf{58.32} & \bf{82.65} \\\hline
\end{tabular}}
\caption{$F_1$ and accuracy results for SC under different OOV settings and candidate generation strategies.}
\label{tab:general_results}
\end{table*}

\subsection{Slot Carryover}

We construct the candidates for the DST as follows: given a set of slot values $W$, we construct another set of slot candidates $\{(w_j, \text{argmax}_{h_i} P(w_j \in h_i \vert \C, \F)), w_j \in W\}$. This last set is then used to train the tracker.

\section{Evaluation}\label{sec:experiments}
\subsection{Candidate generation}\label{subsec:sc_evaluation}

\begin{figure}
\centering
\includegraphics[width=\columnwidth]{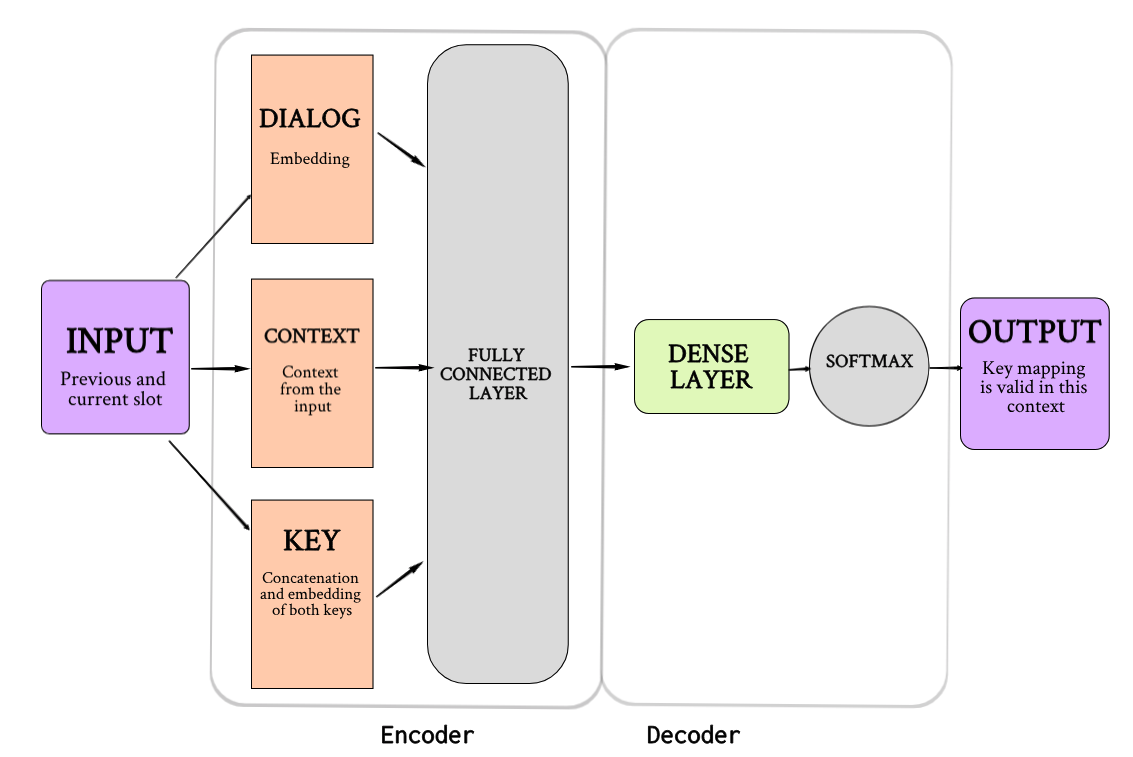}
\caption{MLP: a model that takes in two slots, and determines whether there is a mapping between them. If so, the target slot is passed in to SC for evaluation}
\label{fig:mlp}
\end{figure}

Our experiments are conducted on a subset of data collected from a commercial voice assistant. We train and evaluate the SC model under three candidate generation strategies:

\begin{itemize}
  \item \emph{Hardcoded maps} (HM): The current implementation, which relies on a hand-written, cross-domain, mapping between the keys and their domains to generate slot candidates.
  \item \emph{Multilayer perceptron (MLP)}: The MLP (\figref{mlp}) provides a baseline on the performance of models that do not use KBs. It is trained and evaluated on all the maps from HM, and outputs whether a given map is correct. It achieves 99.99\% test accuracy.
  \item \emph{Our approach}: Our approach constructs a slot candidate for a given context and value. We purposely use a much smaller dataset (about 5\% the size of the data for HM) to train our model, to emphasize the zero-shot conditions.
\end{itemize}

The three mapping strategies are trained in two OOV settings, 0\% and 100\%, where the OOV percentage determines the proportion of keys excluded from the training set. The resulting models are used to classify a corpus from which we generate the training datasets for SC.
For our approach, in high OOV proportions we are performing zero-shot learning, so all the datasets are constrained to be $k=3$ random samples for each of the non-OOV keys. We accept or reject possible keys based on the learned threshold $\tau$.

\subsection{Results}

We train SC in all six datasets, and evaluate it on a fixed test set. We do not retrain the embeddings used by SC. The $F_1$ and accuracy scores across datasets are displayed in \tabref{general_results}. The difference between HM and MLP is negligible, and we can conclude that the MLP learns the mappings properly. This means as well that MLP is bounded by the performance of HM: it is only as good as the underlying probability distribution, and will asymptotically perform the same, regardless of the OOV percentage.
In contrast, our approach assigns a hypernym-based namespace to the data. In the 100\% OOV case, this approach vastly outperforms both strategies. This is likely due to the fact that it is able to inject its own, semantically-related, slot key namespace into SC. This also would explain the lower performance in the 0\% OOV setting.

\section{Conclusions}\label{sec:conclusion}
We presented a model that employs KBs to make better decisions when facing unseen slot keys and values. This approach does not rely on a schema, as it uses the underlying slot key and value attributes, in addition to the given context.
We were also able to learn a distance function for the spaces without making any assumptions other than a hierarchical relation. 
Leveraging KBs means that the vocabulary of this model can be expanded as-needed. This allows its vocabulary to be updated frequently, and without the need to retrain it.

\section*{Acknowledgments}
The authors would like to thank Alyssa Paxevanos-Evans for her help on the early stages of this project.

\bibliographystyle{acl_natbib}
\bibliography{kbs_biblio}

\clearpage
\appendix

\section{Sample Input and Output}

\paragraph{Training input:} As an example, two input triples could be: ("Clint Eastwood", "an American actor, filmmaker, musician, and politician"\footnote{https://en.wikipedia.org/wiki/Clint\_Eastwood},"Actor") and ("Clint Eastwood", "a song by British virtual band Gorillaz"\footnote{https://en.wikipedia.org/wiki/Clint\_Eastwood\_(song)}, "Song"). During inference, the target slot key is not present and the context is leveraged to disambiguate between prospective slot keys.

\paragraph{Constructing the tensors:} Following the example from above, our input slot value will be $w=$ "Clint Eastwood". Assume our KB set is $\mathcal{K} = \{KB_{\text{Personalities}}, KB_{\text{Songs}}, KB_{\text{Companies}}\}$. Our feature tensor $\F$ is then composed of three sets of related terms, each corresponding to a $KB_i \in \mathcal{K}$, as "Clint Eastwood" is a \textbf{personality} commonly identified as an \emph{actor}, \emph{director}, and \emph{composer} for various films, and was \emph{mayor} of Carmel-by-the-Sea, CA. In addition, "Clint Eastwood" is also a \textbf{song} by the \emph{group} Gorillaz that came out in the \emph{year} 2001. On the other hand, no \textbf{companies} exist under such name, and hence the $\F_{\text{Companies},:,}$ entry will be empty.
Note that we might have overlapping values for certain attributes: e.g., $\F_{\text{\textbf{personality}, \emph{actor}, birth date}} = \F_{\text{\textbf{personality}, \emph{mayor}, birth date}}$, which will allow us in turn to discern the features that are relevant for a given slot key.

To construct the hypernym tensor we perform a lookup of all the hypernyms for the associated terms in $\F$, and we include the type of KB. For example, for $\F_{\text{\textbf{songs}}}$, the hypernym vector will be $\H_{\text{\textbf{songs}}} = [\text{song, music, musical composition}]$.

\paragraph{Sample output:} Outputs for our model differ drastically from these of the HM and MLP methods, as it can be seen in \tabref{labels_per_method}. This is mostly due to the fact that the input (target) label $\ell$ is used as a disambiguation guide, rather than an absolute objective, and the output is a concatenation of all the hypernyms found.

\begin{table}[ht]
\centering
\resizebox{\columnwidth}{!}{
\begin{tabular}{l|lll}\toprule
\bf Value & \bf HM	& \bf MLP  & \bf Our approach \\\toprule
"rafi mecartin" & HumanBeing 		& HumanBeing		& WriterDirectorArtist	\\\hline
"ryan allen"	& HumanBeing		& HumanBeing		& ArtistBand \\\hline
"raymond"       & WeatherLocationCity& WeatherLocationCity		& Company	\\\hline
\toprule
\end{tabular}}
\caption{Sample labels emitted by the three different methods employed, for the same value and context. Note how, for the value "raymond", our approach placed too much emphasis on the forklift company, rather than in the geographical location, even when the context (not pictured) clearly requested weather information.}
\label{tab:labels_per_method}
\end{table}

\section{An Analysis of the Bregman Module}
The purpose of the Bregman module is to provide a space-agnostic, learnable metric. We considered more complex approximators to the gradient, such as the Runge-Kutta methods, and variations upon gradient descent \cite{Sutskever2013OnTI}. However, due to heavy calculation times we opted for the symmetric difference quotient approach described in the paper. Other approaches to approximate the Bregman divergence, such as boosting a set of the most commonly used Bregman divergences, were considered and discarded due to their lack of ability to generalize to unseen spaces.

\section{Ablation studies}
We performed ablation studies in both the threshold $\tau$ and in the Bregman module. The inclusion of the threshold increased accuracy by 11\% in our test dataset.
The Bregman module was compared with two simple metrics: squared Euclidean norm and the squared difference. The inclusion of the module increased accuracy by 1\% in our test dataset.

\section{Training Parameters}
We have as trainable parameters the parameters of our Bregman module--with the exception of the step size $\epsilon$--the threshold $\tau$, and the Bi-LSTM parameters.
For the initialization of the Bi-LSTMs, we used Xavier \cite{Glorot2010UnderstandingTD}, and a recurrent dropout of $0.5$\cite{Gal2016ATG}. They have all an input size of $400$, and a hidden size of $200$.
The Bregman module was initalized with random orthonormal matrices \cite{Saxe2013ExactST}, and with an input size of $512$, output size of $1$, and a depth of $10$.
The threshold is initialized to $\tau = 0.75$, and we use Adam \cite{Kingma2014AdamAM} as our optimizer. We trained the model for $50$ epochs, with an early termination of $10$ epochs.
We maintained the maximum number of hypernyms fixed at $9$, the depth of the Wordnet graph at $2$, and the features at $10$.

\end{document}